# Visualizing the Consequences of Evidence in Bayesian Networks

Clifford Champion and Charles Elkan

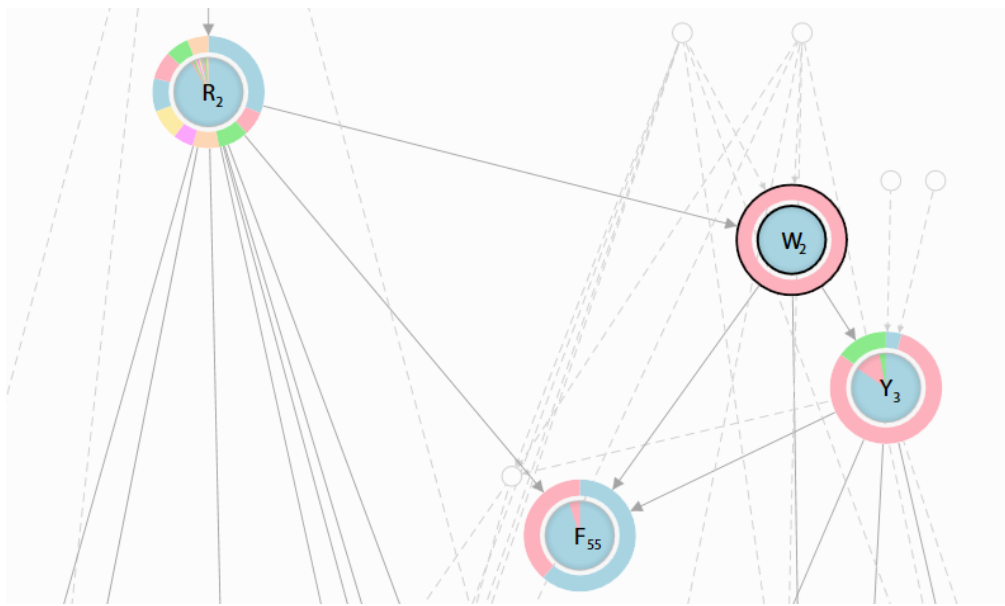

Fig. 1. A relevance-filtered inference diff of two posterior distributions in a Bayesian network of 67 random variables.

**Abstract**—This paper addresses the challenge of viewing and navigating Bayesian networks as their structural size and complexity grow. Starting with a review of the state of the art of visualizing Bayesian networks, an area which has largely been passed over, we improve upon existing visualizations in three ways. First, we apply a disciplined approach to the graphic design of the basic elements of the Bayesian network. Second, we propose a technique for direct, visual comparison of posterior distributions resulting from alternative evidence sets. Third, we leverage a central mathematical tool in information theory, to assist the user in finding variables of interest in the network, and to reduce visual complexity where unimportant. We present our methods applied to two modestly large Bayesian networks constructed from real-world data sets. Results suggest the new techniques can be a useful tool for discovering information flow phenomena, and also for qualitative comparisons of different evidence configurations, especially in large probabilistic networks.

―――――――――――― ♦ ――――――――――――

## 1 INTRODUCTION

The application of machine learning in the 21st century is increasingly both exciting and challenging, with many orders of magnitude more digital data available than before. The amount of text data on the internet has increased from an estimated couple terabytes in 1997, to Twitter.com alone storing 50 gigabytes of new tweets daily [1] [2]. These figures do not count the many private databases used in enterprise, such as the petabytes of customer and transaction information Wal-Mart retains [3].

Though the collection of raw data will continue to have its challenges and costs, significant attention has now turned to the problem of utilizing all of this data. Whether the goal is indexing, data-mining, or building predictive models, today's challenge is fundamentally tied to the enormous number of observations and variables captured. This is the so-called "big data" problem.

Just as important as the algorithms or storage systems are the visualization methods. Presentation is not merely an aesthetic concern. Edward Tufte writes that "often the most effective way to describe, explore, and summarize a set of numbers – even a very large set – is to look at pictures of those numbers," and that data graphics can be both "the simplest [and] most powerful" of methods [4].

Bayesian networks in many applications enable efficient and scalable statistical and causal modeling [5], and have a natural visual representation following from their graph structure. By viewing the graph structure one can quickly identify potential correlations or causal relationships between variables, simply by seeing the presence of edges in the rendered graph structure. Beyond this, more sophisticated analysis by visual means alone is difficult, especially as the model grows in size. Displaying conditional distributions with more than a couple parent variables becomes unwieldy, and networks with upwards of 20 variables can be difficult for a user to navigate and parse visually. Recent work has focused on improving visualization and navigation in large networks of up to thousands of variables [6], and is not a solved problem yet. To understand these large, modern networks more efficiently, and in turn better utilize the wealth of data available in the era of big data, new methods of visualization are needed.

- *Clifford Champion is a graduate of the University of California, San Diego, Computer Science and Engineering. E-mail: cchampio@cs.ucsd.edu.*
- *Charles Elkan is a professor with the University of California, San Diego, Computer Science and Engineering. E-mail: elkan@cs.ucsd.edu*

To this end, we introduce two techniques to assist in the visual analysis of large Bayesian networks: *inference diffs* and *relevance filtering*. After summarizing relevant prior work (section 2), we develop a visual design foundation for our own work (section 3), then define inference diffs (section 4) and relevance filtering (section 5). We conclude with applying these techniques to two real-world data sets (section 6) and parting thoughts (section 7).

## 2  SUMMARY OF PRIOR WORK

Though creating effective ways to visualize Bayesian networks is not a new problem (relative to the age of "Bayesian networks" proposed in detail by Pearl, et al. in 1986 and 1988 [7]), it appears to be a problem that has received relatively little attention. While there have been advances in visualizing large graphs such as those surveyed by Schaeffer [8], these methods depend on basic graph-theoretic information at most, such as cliques and node degree, and don't directly consider the probabilistic aspects of Bayesian networks.

Nevertheless, some variety of visual designs and principles specific to Bayesian networks have been developed or explored over time, and are briefly recounted here.

To visualize causal relationships globally, so-called temporal or causal layouts are popular, placing ancestors (e.g. independent variables) near the top of the visual layout and descendants (e.g. dependent variables) near the bottom, for a generally downward flow of edge directions, for a downward flow of causation. This kind of layout is often used without explicit mention and is a feature of some directed graph layout algorithms, but Zapato-Rivera et al., and Chiang et al. called out this layout explicitly [9] [10].

To visualize local influence (i.e. between exactly two variables) the direction of the edge arrow is of course well-established for indicating the direction of modeled cause and effect. Beyond merely the edge direction, or its presence at all, Zapato-Rivera et al. explored fixed color assignments to independent variables, and color mixtures thereof to dependent variables, weighted to indicate relative strength influence from the parent variables [9]. Zapato-Rivera also considered varying edge lengths, so that mutually influential nodes appear nearer to one another than if uninfluential. Further, both Zapato-Rivera et al. and Koiter explored varying edge thickness to indicate influence between parent and child variables, using thick lines for strong influence. Each discussed various analytical definitions for computing the inputs needed for these visualization techniques [9] [11].

To visualize conditional probability tables (CPTs), Chaing et al. proposed miniature 2d heat maps attached to edges [10], however this is appears to be well-defined only for children with exactly one parent each. Cossalter et al. introduced "bubble lines" connecting nodes in the network to floating CPT windows, making it easier for the user to keep their bearings while debugging CPTs in large networks. They also introduced a numerical difference view for viewing CPTs of two variables expected to have similar local distributions [12].

To visualize the presence of evidence, common practice in literature is to draw a double-border around observed nodes (variables with evidence), or to use shading on the interior of the node. Williams and Amant experimented with using different colors of shading to indicate different evidence values [13].

To visualize marginal and posterior distributions, at least three techniques have been explored. Software application Netica used rectangular nodes instead of circular, in order to embed bar charts for the marginal probability masses of each variable [14]. Software application BayesiaLab allowed the user to open a distribution window for each variable and compare the prior (no evidence) and posterior (with evidence) distributions of a variable in two horizontal bar charts overlaid on one another [15]. Zapato-Rivera et al. used node diameter to indicate large or small posterior probabilities for binary-valued variables, and animation thereof to indicate changes to posteriors under changing evidence [9].

To visualize local and global information simultaneously, Sundararajan et al. employed a partition and fish-eye approach to graph layout, letting the user define and inspect local areas of interest in the network while still seeing the context and structure of the full network [6].

A common trait among most of the approaches is their dependence on relatively static information about the network, whether this be the conditional probability tables, or simple posterior or marginal distributions.

Our goal is to create a visualization that captures a more dynamic view of Bayesian networks, hopefully shedding new light on information flow; and simultaneously to scale effectively in large networks. We outline our basic design choices next, using or iterating on prior work, and upon this foundation introduce a more dynamic approach to visualizing Bayesian networks using inference diffs.

## 3  GRAPHICAL FOUNDATION

### 3.1  Assumptions and Principles

Our approach is to define a visual language equally suited for print or personal computer, and consistent with the principles proposed by Edward Tufte on "graphical displays" [4]. When a human-computer interaction is discussed, we assume one user at a time, that the person is using a mouse or a touch-screen interface, and that the display size and resolution (dots per inch) is that of an average tablet or desktop display.

The underlying model to visualize is a Bayesian network of finitely many random variables, each variable having a finite event space. We assume at the very least the user would desire to be able to view the Bayesian network structure, inspect local conditional probability distributions, see marginal or posterior distributions, inspect the event space of each variable, and otherwise clearly see the basic makeup of the network instance. These assumptions are sufficient for defining the foundation of our graphic design.

We also seek to avoid distorting data or potentially misleading users, and to avoid unnecessary ink, all-in-all minimizing so-called chartjunk and maximizing the so-called data-ink ratio [4]. Every pixel or drop of ink should convey information and convey so unambiguously.

### 3.2  Network Structure and Random Variables

The most basic information in a Bayesian network is the structure and random variables therein. To present the structure is to present the (learned or constructed) causal influence between variables. Logically, this object is a directed acyclic graph (DAG); visually, this is traditionally a collection of labeled circles and arrows between them. We largely continue this tradition.

Random variables must be clearly identifiable, while at the same time clutter must be kept under control, otherwise it becomes noise. To this end, there are two views: structural and legend. The structural view presents each variable as a single, circumscribed capital letter, taken from the first letter in the name of the variable, visible in Fig. 3. In the legend view, each letter is then mapped to its full variable name, such 'A' to 'age'. Capital letters are used in the structural view for readability. As with previous methods, vertical ordering in the structural view is presented causally top-down to the extent possible, though this is not always possible, especially in large networks.

Fig. 2. Values in each random variable event space are assigned a color in the legend view using a predetermined color palette.

To scale to large networks, we perform two more things. First, where two or more random variables have the same single letter in the structural view, we suffix their name with a unique number, chosen sequentially from 1. This numerical suffix appears both in the structural view and the legend view in subscript type. Second, both views are scrollable and both follow loosely the same top-to-bottom variable ordering.

On the appearance of random variables in the structural view, we de-emphasize the dark stroke that traditionally circumscribes the variable, as we will be using this stroke to carry meaning later. Until then, this shading decision comes at cost our to data-ink ratio.

Finally, we reserve the use of color when only presenting structure, so that color may be used to present other aspects of the model, discussed next.

### 3.3 Coloring the Event Space

Representing values from the event spaces of our random variables clearly and succinctly to the user will be critical for inference diffs discussed later. We adopt the use of color here, assigning for each possible value in each event space a unique color. We do not guarantee unique colors over all event spaces in the model: avoiding the constraint permits us to design a reusable, optimal color palette with minimal visual confusion for single variables at a time. Though there is possible ambiguity in values from different spaces sharing a same color, our design minimizes this ambiguity by always framing color in the context of a specific variable. Where two or more variables share the same event space, we reuse the color mapping for consistency and reinforcement.

To indicate what each color assignment is, we augment our legend view to list each event space value and corresponding color, seen in Fig. 2.

For categorical event spaces our color map is constructed so that no two contiguous colors are perceived too near to one another: for example, orange may follow green but may not follow brown. For ordered event spaces the color map is constructed in the opposite fashion by sequentially choosing neighboring hues on the color wheel, e.g. from the blue region, through yellow, to the red region.

As will be important later, we ensure any presented color order (value order) is constant, e.g. for a particular event space, blue always appears first before orange.

### 3.4 Conditional Probability Tables

One of practical difficulties with Bayesian networks can be the size of each conditional probability table (CPT) or its local distribution. The size of a random variable's CPT is generally the number of probability weight assignments specified, which grows exponentially in the number of parent variables (or the in-degree of the variable). Some tools present the CPT as a single table with columns for each permutation of parent values, but this tends to require a large horizontal scrolling area. We propose that vertical scrolling is more natural, and present our CPT vertically when shown. To do this we present each conditional probability distribution for a given parent permutation as a vertical list of probability densities. We stack each such list vertically, and separate each by their corresponding parent value permutation, identified above it. We use the event space color mappings here, for each probability density and each parent value; and again use corresponding abbreviated variable names (e.g. A$_1$).

### 3.5 Embedded Distributions

Viewing the marginal distribution of each variable should be convenient; that is, we want a clear way to see $P(X)$ for each random variable $X$. Most tools require the user open an additional window to see such distributions, either in tabular form or bar chart. We embed the distribution directly in the variable in the structural view as in Fig. 3. To do this we construct a pie chart using our event space color mapping, and render pie slices proportional in size to the posterior probability mass of each value, starting at the 12 o'clock position, allocating slices in clockwise order.

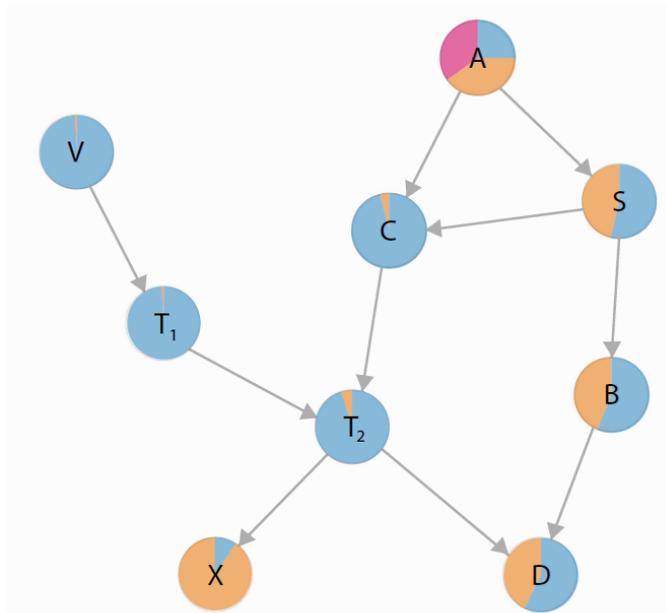

Fig. 3. Marginal distributions are embedded directly in the variable via area pie chart.

This highly visual approach conveniently presents an overview of the entire statistical model to the user, without their needing to inspect variables one-by-one in sequence or in additional views. Where more precise numerical inspection is needed, we show an additional color-coded tabular view similar to our CPT view.

### 3.6 Evidence

Arguably the greatest power of a Bayesian network model is in computing posterior distributions of arbitrary evidence, i.e. $P(X|\boldsymbol{E})$.

From a visualization perspective it is important that the user clearly see which variables currently have evidence and which do not; and furthermore, what specific values of evidence have been specified. To indicate that a variable has user-defined evidence, we circumscribe the variable in the structural view with a strong black stroke. Moreover, the interior of node is colored entirely with the associated color of the evidence value as in Fig. 4.

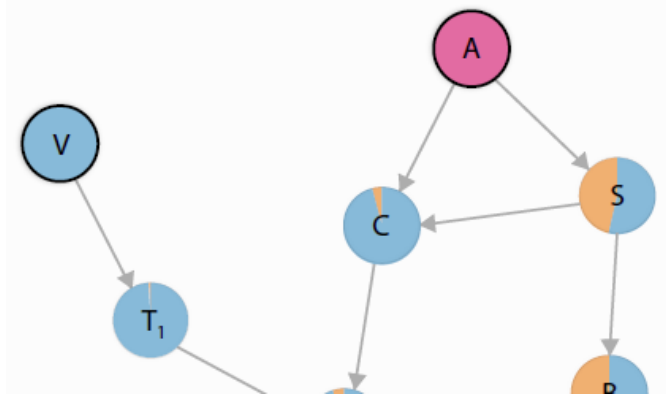

Fig. 4. Evidence nodes are circumscribed with a strong black stroke, and colored according to their evidence. All other nodes' embedded distributions are updated to reflect their (marginal) posterior distribution. For example, $T_1$'s embedded distribution now reflects $P(T_1|V = v, A = a)$ rather than just $P(T_1)$.

Finally, all embedded distributions for non-evidence nodes are updated in the structural view to reflect each variable's new, posterior distribution. Our previous visualization which embeds marginal distributions is a special case of visualizing posterior distributions with an empty evidence set. We will formalize our notion of evidence in finer detail shortly.

# 4 INFERENCE DIFF

## 4.1 Motivation

With a visual foundation established, we focus our attention to more sophisticated visual analysis methods. The novel idea we introduce first is that of an *inference diff*.

Inference and information flow is an important capability of Bayesian networks. Consider a large network which models the health of components in a large multi-component system. One may wish to use this network to ask which components' probabilities of failure are affected by one or more other variables, for instance ambient air temperature, and for those affected, to what degree; or the inverse of this and ask what are the most likely environmental conditions given a failure in one or more components. As networks grow in size, the answers to these questions can be as difficult to find as the right question to ask in the first place.

There are analytical tools such as d-separation; however, such a tool is limited in its application, largely because a Bayesian network is itself inherently limited in its ability to describe certain independent relations [16]. Recall that for a network $G$, its I-map $I(G)$ may not be a minimal I-map, meaning $G$ contains unnecessary edges and is too safe in its conditional independence claims. Moreover, for a true joint distribution it may be impossible for any Bayesian network to have a perfect I-map (or P-map). There may also be context-specific independencies in the network, not discernible in network structure alone. Further, the user may not need to know or care about certain dependencies if small or approximately independent. In each case these issues can lessen the usefulness of d-separation analysis in practice, or require more complicated analysis.

On the other hand there is exhaustive computation, using inference algorithms to produce complete posterior distributions for some or all of the non-evidence variables. Such output is much more detailed and precise, but suffers from the limitation that it is static. From it, there is no direct indication of how or to what degree belief propagation occurred. There is only a before state and an after state.

What we would like is a way to visualize in an obvious way the effects of information flow through the network. For this we find inspiration in modern software engineering practices and so-called "diff" tools (short for "difference" tool). Reviewing the "diff" of two human-readable files is an every-day practice in commercial software development, generally aided by the use of color and side-by-side before-and-after views. As a means of visualizing change, diffs are highly efficient and visually intuitive. Our goal is to find an as-effective method of viewing inference and information flow in Bayesian networks, in hopes of enabling a more powerful kind of visual analysis.

## 4.2 Definition

We start with a mathematical definition an inference diff. Given a Bayesian network $B$ describing a probabilistic model over $n$ random variables $\{X_i : i \in [1,n]\}$, each with finite event space $S_i$, and given two evidence sets $\boldsymbol{E_1}$ and $\boldsymbol{E_2}$, each an element of the set of partial observations $\times_{i=1}^{n} S_i \cup \{?\}$, we define an *inference diff* $\Delta$ as the set of pairs

$$\Delta = \{(P(X_i|\boldsymbol{E_1}), P(X_i|\boldsymbol{E_2})) \; : \; i \in [1,n]\}$$

where element '?' indicates an unobserved or unspecified value.

In other words, an inference diff is the set of pairs of conditional probability distributions, for each random variable, and according to two sets of evidence.

For example if the random variables of the network are $X$, $Y$, and $Z$, and $Z$ takes on value $z$ in $\boldsymbol{E_1}$ then $\boldsymbol{E_1} = (?,?,z)$ and $P(X|\boldsymbol{E_1}) = P(X|Z=z)$. If an evidence set is equal to $(?,?,...,?)$ we say that it is empty.

## 4.3 Visualization

To visualize inference diffs we extend our use of pie charts. First we establish that evidence set $\boldsymbol{E_1}$ is in fact the evidence set when using the network to view a single set of posterior results. For instance Fig. 4 represents an $\boldsymbol{E_1}$ with values for two variables, and an $\boldsymbol{E_2}$ that is empty. To augment our visualization for the case when $\boldsymbol{E_2}$ is non-empty, we introduce for each variable a "ring" chart, concentric with the variable's existing pie chart. We reuse the event space color map established for that variable's event space, maintain a consistent event space ordering, and again weight the slices in proportion to posterior probability masses for that variable, this time conditioned on $\boldsymbol{E_2}$.

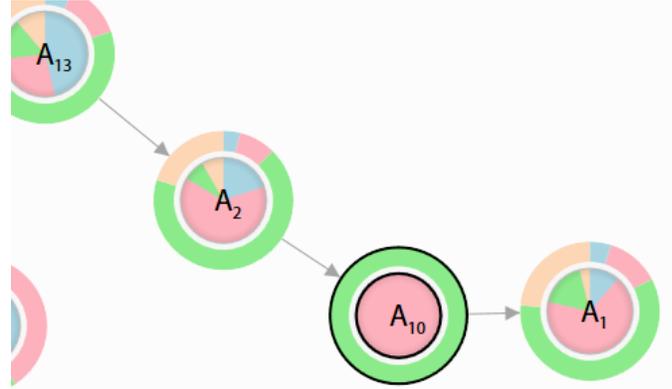

Fig. 5. An inference diff between two evidence sets. Variable $A_{10}$ has observed values in both sets, indicated by a black stroke around both the inner circle (evidence set 1) and the outer ring (evidence set 2).

To indicate which variables have evidence specified, we continue to use the strong black stroke, applied to either the pie, the ring, or both, in accordance with which variable and evidence set has evidence. By reserving use of the black stroke earlier, we are able to apply it here in a more nuanced fashion, to help disambiguate from which evidence set a variable's evidence is specified.

This concentric design allows the user to make direct comparisons of the effects of evidence sets $\boldsymbol{E_1}$ and $\boldsymbol{E_2}$, quickly and easily. At least two classes of queries can be performed now and produce interesting visual answers. First, one can view information flow concretely, by setting $\boldsymbol{E_1} = (?,?,...,?)$, and to $\boldsymbol{E_2}$ any other partial observation. Second, one can make direct comparisons between different non-empty evidence sets, such as asking whether observing some three variables is different than observing only two of them, or asking how different is it to observe variable $X_i$ having some value than to observe variable $X_j : j \neq i$ having that value.

# 5 RELEVANCE FILTERING

While the inference diff enables direct comparisons of the effects of different evidence on each variable, it doesn't necessarily help guide the user to the variables they may be interested in the most. This especially can be a problem in large Bayesian networks, where there is too much information visible simultaneously, or where the user himself lacks familiarity with the model. What we would hope to achieve is a way to guide the user to the variables they are likely to be interested in, given some provided evidence. To accomplish this we use the inference diff as our basis, and add to it *relevance filtering*.

## 5.1 Definition

We define the relevance of a random variable $X_i$ as the symmetric Kullback-Leibler (KL) divergence [17] of that random variable given its inference diff. More precisely, given an inference diff $\Delta$ derived

from evidence sets $E_1$ and $E_2$, we define the relevance of a random variable $X_i$ as

$$r_\Delta(X_i) = D_{KL}\big(P(X_i|E_1) \parallel P(X|E_2)\big) + D_{KL}\big(P(X_i|E_2) \parallel P(X|E_1)\big).$$

The function $D_{KL}$ indicates the Kullback–Leibler divergence, with standard definition

$$D_{KL}(P \parallel Q) = \sum_i \ln\frac{P(i)}{Q(i)} * P(i)$$

for probability distributions $P$ and $Q$ sharing a single event space. We use the symmetric divergence so that our definition of relevance is also symmetric.

Given two random variables $X_i$ and $X_j, i \neq j$, if $r_\Delta(X_i) < r_\Delta(X_j)$ then we say that $X_j$ is *more relevant* than $X_i$ given evidence sets $E_1$ and $E_2$. This definition of relevance is chosen intuitively. Because distributions which differ greatly have high KL divergence values, we are saying that the variables whose conditional distributions changed the most (between $E_1$ and $E_2$) are the variables that are most relevant to the user.

### 5.2 Visualization

With relevance defined, we have our final mathematical tool to complete the visualization method. For large networks it can be easy for arbitrary evidence sets to produce negligible differences in the posterior distributions of many variables. It is with this situation in mind that we apply the definition of relevance.

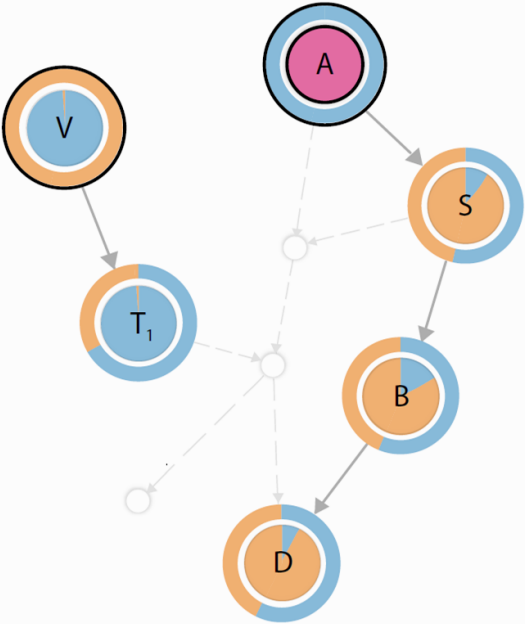

Fig. 6. An inference diff with relevance filtering enabled. Variables $A$ and $V$ contain evidence in at least one evidence set each. Compared with the structure seen in Fig. 3, variables labeled $C$, $T_2$, and $X$ were least relevant given the evidence sets, and thus are reduced in size and visibility.

To do this, first we decide which variables are relevant enough for the user: given an inference diff $\Delta$, we sort the variables of $\Delta$ in descending order according to their relevance. Second, we introduce a user-configurable relevance threshold, represented as a percent value $c$ between 0% and 100%. Finally, we decide for each random variable $X_i$ whether it is "relevant" or "irrelevant" according to whether $X_i$ is in the top $c$ percent of variables ordered by relevance.

Finally, we adjust the visualization in the structure and legend views. Variables which are irrelevant according to threshold $c$ are shrunk, dimmed, and their pie and ring charts removed in the structure view. We also shorten the edge lengths between any two collapsed variables, and edges connecting at least one irrelevant variable are changed to a dotted line rendering.

The overall effect is to shrink the virtual space needed for the entire graph structure, which focuses the user's attention on the relatively small number of variables remaining. For these remaining variables we continue to show the pie and ring charts associated with the inference diff. We also remove from the legend view irrelevant variables.

The end result is a user-controllable level of visual complexity, and a clear, concise, and qualitative view of where information flow has most impact.

## 6 APPLICATION TO REAL-WORLD DATASETS

To test the usefulness of the visualization techniques described above, we developed a software application called *B-Vis*. In building this application we implemented our own structure learning and inference algorithms based on existing literature. We have open-sourced the application and learning modules as part of *F-AI* [18]. For graph layout we utilized the library *Graph#* [19], making small modifications as necessary. Unless otherwise noted we use the "Sugiyama Efficient" method found in *Graph#*, which is presumably based on one or more of the layered graph drawing algorithms of Kozo Sugiyama [20].

### 6.1 San Francisco Traffic Data Set

Our first data set consists of San Francisco bay area highway system traffic flow measurements ("Traffic" set), acquired by Krause and Guestrin [21], and modified by Shahaf et al. [22] into four discrete bins of traffic flow quantity. The set consists of 4,415 samples and 32 traffic sensor measurements per sample with no missing data. From this we train a Bayesian network, learning both structure and CPTs, using a uniform Dirichlet prior and an in-degree limit of two for each variable.

Fig. 7 shows the learned structure, final layout, and embedded marginal distributions. Note that the network size and structure makes it inherently tall, such that the view must be zoomed out in order to fit the entire network on a standard resolution display without scrolling.

Next we configure some evidence. We let $E_1$ be empty and $E_2$ contain an evidence value of 'medium' (mapped to green) for variable $A_4$. We configure relevance filtering to preserve the top 20% most relevant variables. Fig. 8 shows our visualized result.

Note that because of relevance filtering, not only do we clearly see the variables which are likely to interest us the most, but we are also able to zoom in on our network view, fitting all of the network on-screen without scrolling; embedded posterior distributions on the relevant variables are clear to see and compare. We chose variable $A_{11}$ for this example because of its potentially large influence according to the network structure, with a vertex degree of 14. Relevance filtering in this instance presented a visually manageable total of only 5 connected variables, those 5 with which were most affected by the change in evidence. Because the Sugiyama layout algorithm is layered there is also a resemblance between the before and after layouts.

The Traffic data set is somewhat of a special case in that all variables share the same event space, and that each of the random variables are highly correlated for identical values in the event space.

### 6.2 U.S. 1990 Census Data Set

For a larger and more interesting network, we consider the 1990 U.S. Census data set ("Census"), discretized by Meek et al. [23] [24]. This set contains responses and classifications of 2.4 million individuals stored in 67 attributes. For the purposes of testing our visualizations, work we train on only the first 10,000 randomly chosen examples. As with before we use a uniform Dirichlet prior, but increase our in-degree limit to 3 for this training.

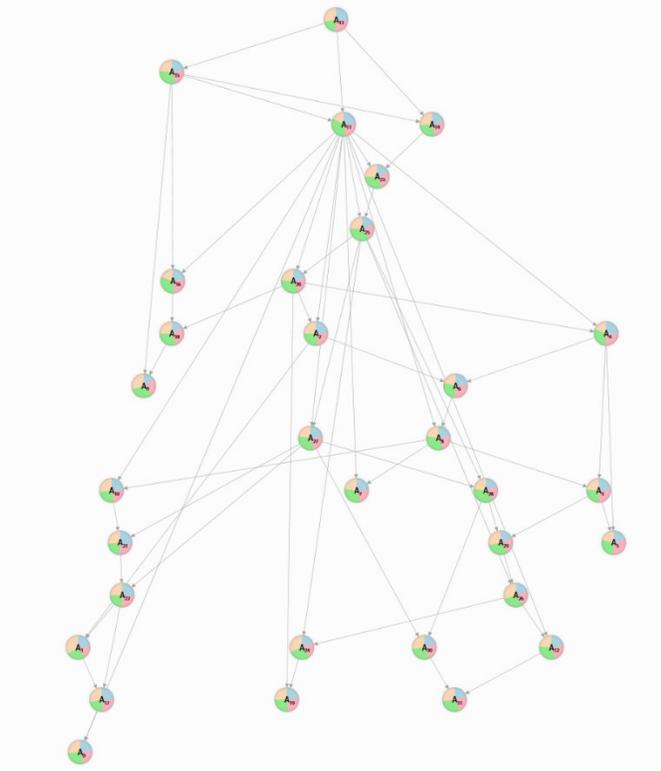

Fig. 7. A network trained on the Traffic set. No evidence sets are specified, revealing all variables and their marginal distributions.

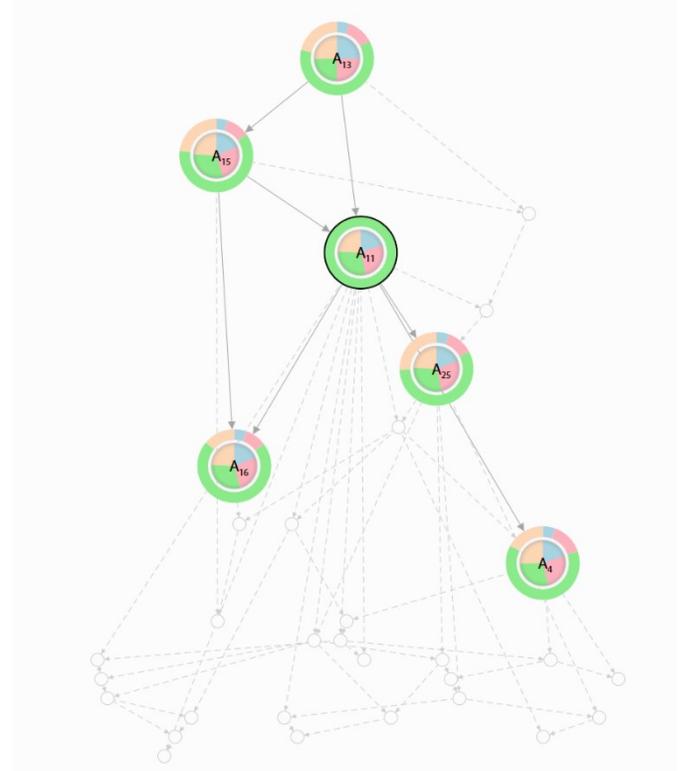

Fig. 8. An inference diff of the Traffic network with relevance filtering enabled, choosing the top 20% most relevant variables given the evidence sets.

The trained network is visible in its entirety in Fig. 9. We again use the Sugiyama layout method. Note that due to the size of this network, we must zoom out quite far to see it in its entirety.

To test the usefulness of relevance filtering we configure evidence set 1 in our inference diff to be empty, and evidence set 2 an observation of 'true' for variable 'income4'. Variable 'income4' indicates that the individual reported some form of interest, dividend, or rental income/loss in the prior year [23] [24] [25]. Most individuals in the U.S. report no income of this kind, so it is an interesting question to ask the network model: how does knowing a person receives income of this type change our other statistical beliefs about the person?

Though the inference diff provides direct before-and-after comparisons, sixty-six other variables is a large number to sift through. We configure relevance filtering to again return the 20% most relevant variables, and find some interesting results, both as it pertains to the model, and to the visualization, seen in Fig. 10.

The visual analysis quickly reveals that the twelve variables most affected (in no special order) are: year of immigration ('immigr'), place of birth ('pob'), their Hispanic heritage ('hispanic'), the relationship to the homeowner ('relat2'), whether the individual is part of a subfamily ('subfam1'), how many subfamilies are present ('subfam2'), whether the individual works on a farm ('income3'), whether the individual served in the military during no major war or conflict ('othrserv'), their ancestry ('ancstry2'), their means of transportation to work ('means'), their status in the job market ('avail'), and the employment status of their parents ('remplpar') [25].

With the twelve (or generally, configurable number of) most interesting variables brought to the user's attention, they can now drill down into more detail. For instance the updated belief about means of travel to work has changed in an interesting way. Fig. 11 shows the inference diff for variable 'means'. We see that knowing the person receives interest, dividends, or rental income greatly increases the statistical chance that they do not commute to work each day, and greatly reduces the chance that they commute by means other than car, such as bike or rail.

Another feature of the output is revealing. Note that none of the three variables directly connected (structurally) to 'income4' were included in the filtered output ('rearning', 'rpincome', and 'income6'). Though this may seem counter-intuitive, it is explicable. Depending on the local distributions (e.g. CPTs) at each variable, the effects of evidence can amplify from link to link according to inference, in a manner similar to error accumulation. Further, with three trails out of 'income4', there is opportunity for small local changes to join elsewhere with a combining effect. Or the opposite can occur, such as with the example evidence sets used on the Traffic network (Fig. 8). Thus these visualization methods can be revealing of certain information flow behaviors in Bayesian networks, in general and for specific evidence sets.

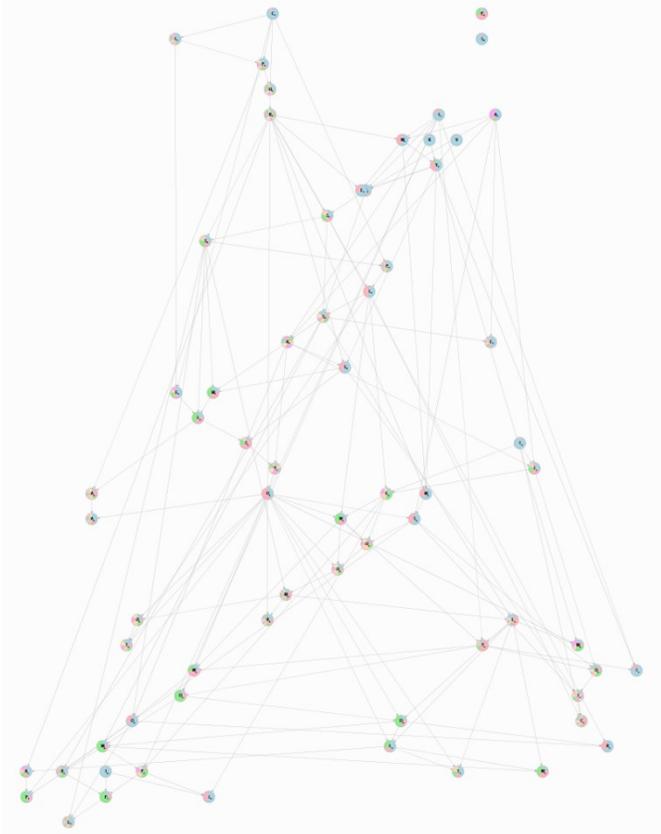

Fig. 9. Census network, consisting of 67 variables, zoomed out to reveal the entire structure.

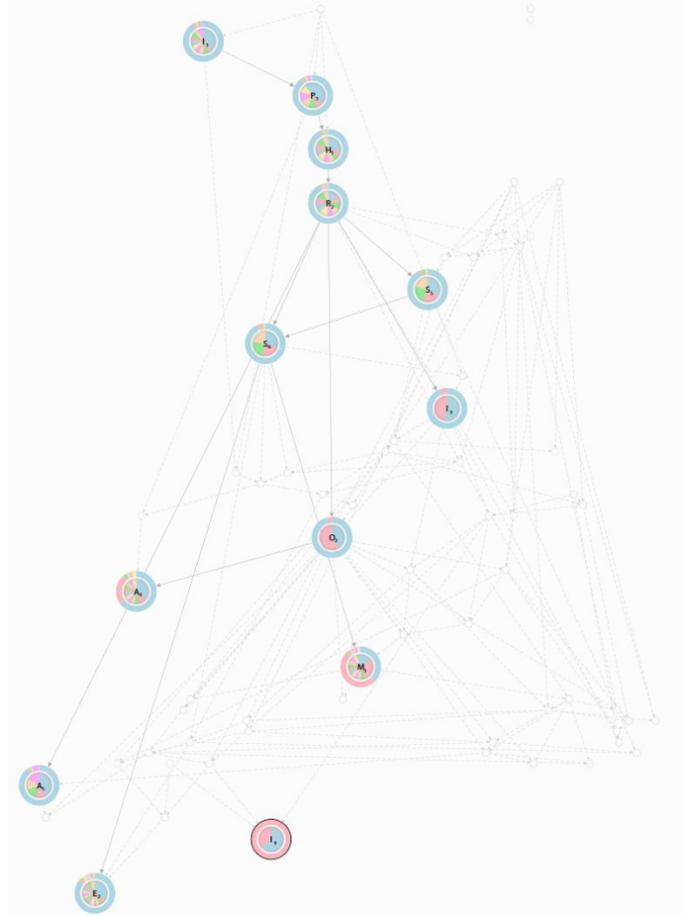

Fig. 10. Census network, with relevance filtering enabled for the current inference diff. The top 20% most relevant variables retain their embedded posterior distributions, while all other variables are reduced in size and visibility.

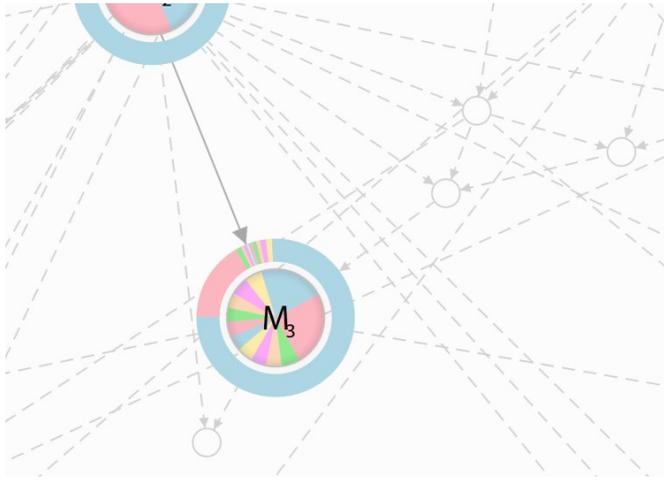

Fig. 11. An inference diff of the variable 'means' (of transportation). The diff is generated from an empty $E_1$, and an $E_2$ with $'income4' = true$.

## 7 FUTURE WORK

### 7.1 Challenges and Scaling Further

Maintaining layout stability is particularly important. The user must be able to adjust relevance filtering without radical changes to the layout, otherwise the experience quickly becomes disorienting. We were able to maintain a sufficient level of stability by using a Sugiyama-based layout. Though this has inherent stability due to its layered approach, it is not perfect for our needs. We would like to incorporate a customized or more sophisticated layout algorithm with inference diffs and relevance filtering in mind. Such an algorithm may continue to be layered, for instance with stability addressed in more detail in Sugiyama's original works [20]; or forced-directed with constraints, which has ongoing exploration [26]. It may also be possible to enhance the fish-eye techniques of Sundararajan et al., by automatically configuring their interest areas using the locations of relevant variables found using relevance filtering [6].

The choice of color as a modality for values is problematic when scaling to large event spaces. For instance, some variables in the Census data set contained over 15 possible values. Our color mapping at present contains only 6 unique color values, meaning that for such variables some colors were used multiple times. Because we present a consistent ordering, both radially and in the legend view values, ambiguity is mostly removed, but requires additional mental energy on the part of the user. For categorical variables with significantly more than 15 values, the effectiveness of the color mapped approach is expected to fall apart. For color-blind users, using a limited color palette creates further difficulty in scaling. One interesting possibility may lay in collapsing colors in inference diffs, where the probability masses of certain values are small or have changed very little in the diff for that variable. Coloring continuous-valued variables is not addressed here, but may be possible as well, perhaps by bounding the event space and assigning distinct colors to special points of significance in the event space, with weighted color blending for intermediate values.

### 7.2 Applications to Other Types of Graphical Model

With respect to Bayesian networks, though the trained models presented above are statistical rather than causal, an inference diff is possibly most useful in a network with true causal modeling. For example in medicine, one could quickly ask the network for a visual answer to the question: given a patient with conditions $X = true, Y = true$ in $E_1$ and $E_2$, what are the largest differences expected between prescribing treatment $A$ and treatment $B$, i.e. between $do(A = false, B = true)$ also in $E_1$, and $do(A = true, B = false)$ also in $E_2$.

Other probabilistic network models may benefit from inference diff and relevance filtering visualizations, such as Markov random fields which have similar inference and belief propagation capabilities.

### 7.3 Unused Modalities

Unused visual modalities could be explored, or reincorporated from prior work. Edge thickness and edge colors could possibly carry additional information. Having more than two evidence sets, by using more than one ring chart per variable, may also be possible and useful. Instead of beginning embedded distributions' slices at the 12 o'clock position, the start degree could be varied to carry significance of some kind. The geometric shape of nodes in the structural view could also convey information, such as whether the node captures a causal dependency.

### 7.4 Other Metrics for Relevance

Metrics other than KL-divergence may more appropriate for more specialized relevance filtering. Regardless of the metric, it may also be useful if user-defined weights can be attached to values in the event spaces, such that large changes to probability masses of low-weight events account for less in making that variable relevant.

## 8 CONCLUSION

Visualization methods are increasingly important as the scope and quantity of data increases. The flexibility and distributed nature of Bayesian networks, and graphical models in general, make them one of many useful tools in machine learning. In this paper we introduced and refined a new set of visualization principles and techniques.

First, we contend that viewing the structure of a network is but half the visual story. We complete it by placing posterior distributions directly in the network as embedded pie charts. Second, direct comparisons are incredibly useful tools, yet past visualizations have struggled with this. We propose inference diffs as a method of meaningful direct comparisons, using concentric pie and ring charts for visualization. And finally, navigation in large Bayesian networks is generally a challenge. We introduce relevance filtering, with KL divergence as a mathematical basis, as a tool to guide the user to variables of interest in the model.

### ACKNOWLEDGMENTS

The authors wish to thank the faculty at the CSE department at U.C. San Diego.